\documentclass{article}
\usepackage{spconf,amsmath,graphicx}
\usepackage{amssymb,bm}
\usepackage{url}

\usepackage[ruled,vlined]{algorithm2e}

\title{Error Diffusion Halftoning Against Adversarial Examples}
\name{Shao-Yuan Lo \hspace{0.1cm} and \hspace{0.1cm} Vishal M. Patel \thanks{This work was supported by the DARPA GARD Program HR001119S0026-GARD-FP-052.}}
\address{Dept. of Electrical and Computer Engineering, Johns Hopkins University, MD, USA \\\texttt{\{sylo, vpatel36\}@jhu.edu}}

\begin{document}
%\ninept
%
\maketitle
\begin{abstract}
Adversarial examples contain carefully crafted perturbations that can fool deep neural networks (DNNs) into making wrong predictions. Enhancing the adversarial robustness of DNNs has gained considerable interest in recent years. Although image transformation-based defenses were widely considered at an earlier time, most of them have been defeated by adaptive attacks. In this paper, we propose a new image transformation defense based on error diffusion halftoning, and combine it with adversarial training to defend against adversarial examples. Error diffusion halftoning projects an image into a 1-bit space and diffuses quantization error to neighboring pixels. This process can remove adversarial perturbations from a given image while maintaining acceptable image quality in the meantime in favor of recognition. Experimental results demonstrate that the proposed method is able to improve adversarial robustness even under advanced adaptive attacks, while most of the other image transformation-based defenses do not. We show that a proper image transformation can still be an effective defense approach. Code: \url{https://github.com/shaoyuanlo/Halftoning-Defense}
\end{abstract}

\section{Introduction}

The great success of computer vision methods over the recent years can be largely attributed to deep neural networks (DNNs). However, DNNs are vulnerable to adversarial attacks, in which carefully crafted imperceptible noise can mislead DNNs into misclassifying input data \cite{Szegedy2014Intriguing,goodfellow2015explaining}. Hence, reliable defense mechanisms are critical for achieving robust computer vision systems. The existing empirical defense approaches can be categorized into two streams. The first one is based on adversarial training \cite{madry2018towards}, which generates adversarial examples on-the-fly during the training process to train a model, making the model learn robust features for recognition. This stream has been repeatedly validated as effective, especially under strong adaptive attacks in the challenging white-box setting \cite{obfuscated}. It has been widely used as a fundamental defense backbone \cite{Xie_2019_CVPR,Wu2020Defending,shao2020open}.

The other stream deploys an image transformation at the pre-processing stage before inference to protect DNNs from adversarial effects. The intuition is that such image pre-processing could filter out adversarial perturbations, allowing one to feed an adversary-free image to a classifier. Specifically, consider a target classifier $C$, an adversarial example $x_{adv}$ and its ground-truth label $y$, where $C(x_{adv}) \ne y$.   In these approaches, the idea is to find a transform $T$ such that $C(T(x_{adv})) = y$. Many types of transformations have been adopted in the context of adversarial defense. For example, bit-depth reduction quantizes pixel values to invalidate adversarial variations in an image \cite{xu2017feature,guo2017countering}. JPEG compression performs quantization in the frequency domain to remove perturbations \cite{dziugaite2016study,das2018shield,jia2019comdefend}. Image denoising operations such as mean filter, median filter and non-local means \cite{buades2005non} have been used as defenses as well \cite{Xie_2019_CVPR,li2017adversarial}. However, almost all of these attempts have been defeated under the white-box threat model \cite{obfuscated,carlini2017adversarial}. That is, if attackers are aware of the presence of the defense, they are able to incorporate the defense into adversary search or combat obfuscated gradients. Therefore, this defense stream is declining. Raff et al. \cite{raff2019barrage} stochastically combined a lot of image transformations to defend against the adaptive attacks, but failed to maintain the performance on clean images. The clean data performance is largely sacrificed owing to the multiple transforms. 

\begin{figure}[t!]
	\begin{center}
		\includegraphics[width=0.34\textwidth]{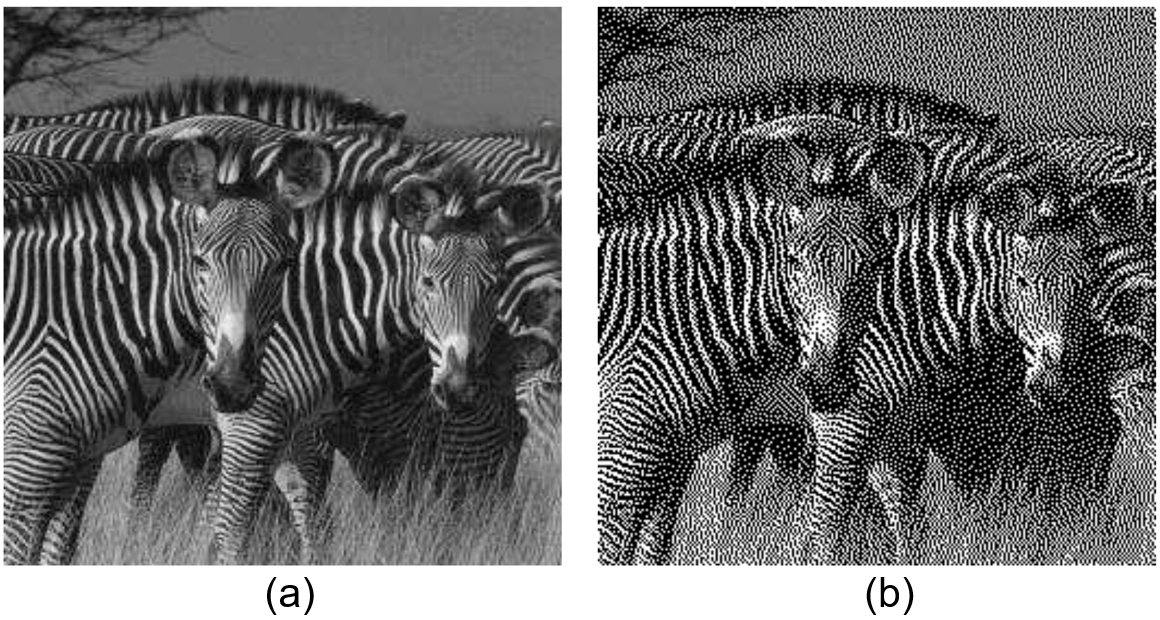}
		\vskip-13pt\caption{(a) Original image. (b) Floyd-Steinberg halftone.  One can clearly see the geometric structure present in the halftone.}
		\label{fig:halftone}
	\end{center}
\end{figure}

In this paper, we propose a novel defense method based on error diffusion halftoning \cite{floyd1976an,jarvis1976survey}. Different from most of the other image transformation-based defenses, it mounts resistance to adversarial examples even after accounting for the challenging adaptive attacks. To the best of our knowledge, this work is the first to leverage digital halftoning as the transformation for adversarial defense purposes. 

Digital halftoning, which sometimes referred to as spatial dithering, is a process of rendering a grayscale image into a binary image (i.e., black-and-white) 
\cite{ulichney1987digital,funkhouser2000image,bouman2020digital,easley2009inverse}. There are several commonly used halftoning algorithms. Thresholding quantizes each pixel value by comparing it with a fixed threshold. It is the simplest algorithm but results in poor rendering quality. Random dithering compares each pixel value with a random threshold to randomize quantization errors. This somewhat remedies the downside of thresholding. Ordered dithering creates a dither matrix to turn the pixels black or white in a specified order, yielding better halftoning results. Error diffusion dithering \cite{floyd1976an,jarvis1976survey} contains an error diffusion mechanism that disperses quantization errors to neighboring pixels. This belongs to an area operation rather than a simple pointwise operation and can mitigate visual artifacts.  We choose error diffusion dithering as our halftoning transformation and deploy it at the pre-processing stage for defense. Specifically, Floyd-Steinberg dithering \cite{floyd1976an} is used in our approach.  Fig. \ref{fig:halftone} shows a sample image and the corresponding Floyd-Steinberg halftone.  As can be seen from this figure, even though the halftone only consists of black and white dots, it maintains the overall structure of the object present in the image.  

Error diffusion halftoning quantizes pixel values to filter out adversarial perturbation, and the error diffusion mechanism can weaken adaptive attacks. Moreover, spreading quantization errors produces higher halftoning quality and thus maintains better accuracy on clean data. The proposed method significantly improves robustness against different adversarial attacks, including PGD \cite{madry2018towards} and MultAV \cite{lo2020multav}, under the white-box setting. In the meantime, it is able to achieve good clean data performance. Although most of the image transformation-based defenses have been proved ineffective, we show that this stream is still worthy to explore.

\section{Proposed Method}

\begin{figure}
	\begin{center}
		\includegraphics[width=0.25\textwidth]{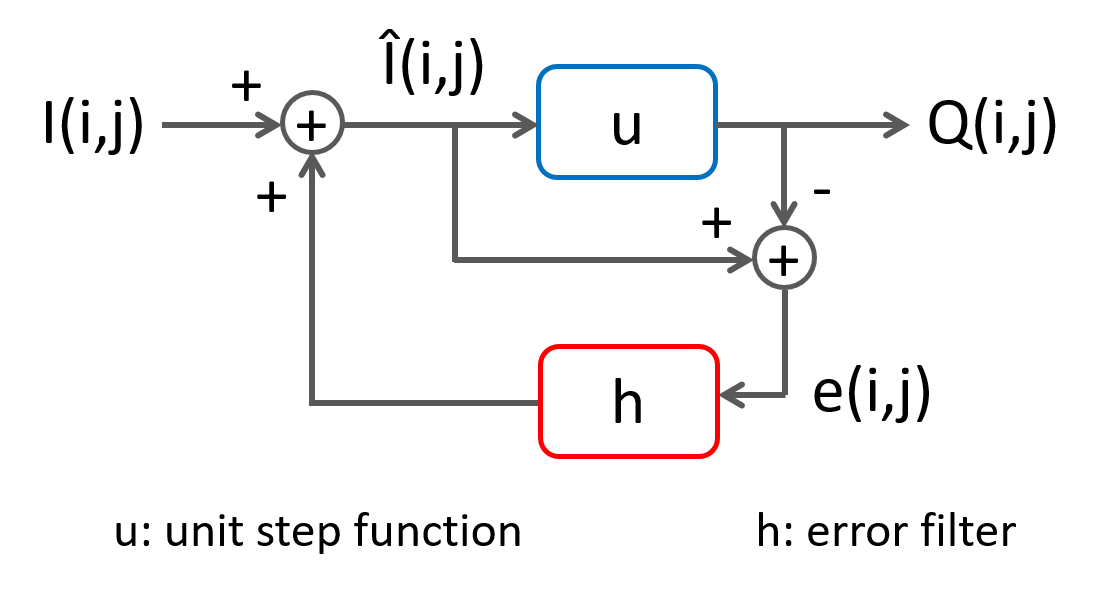}
		\vskip-13pt\caption{Error diffusion halftoning scheme.}
		\label{fig:diagram}
	\end{center}
\end{figure}

We use error diffusion halftoning \cite{floyd1976an,jarvis1976survey} as the image transformation for the adversarial defense purpose. The key idea is to quantize each pixel in the raster order (from left to right, top to bottom) one-by-one, and spread the quantization error to the neighboring pixels. Beginning with the top-left pixel, the pixel value is binarized by thresholding, then the quantization error is dispersed to neighboring pixels using pre-defined weights. Following the raster-scan indexing scheme, the procedure continues until the bottom-right pixel has been transformed. More precisely, let us consider an input image $I$ with pixel values $\in [0,1]$, and an error filter $h$. For each pixel $I(i,j)$ in $I$ with the raster order, it pulls error forward as:
% pull error forward
\begin{equation}
\label{pull}
\hat{I}(i,j) = I(i,j) + \sum_{m,n \in S} h(m,n) e(i-m,j-n).
\end{equation}
Next, $\hat{I}(i,j)$ is quantized to a binary value:
% quantization
\begin{equation}
\label{quantization}
Q(i,j) = u(\hat{I}(i,j) - \theta),
\end{equation}
where $u()$ is a unit step function with a threshold $\theta = 0.5$. The pixel's quantization error is calculated as: 
% error
\begin{equation}
\label{error}
e(i,j) = \hat{I}(i,j) - Q(i,j).
\end{equation}
Then, this error is pushed ahead, and the next pixel in the raster order pulls the errors, repeating from Eq. (\ref{pull}) to Eq. (\ref{error}) until the last pixel. Fig. \ref{fig:diagram} summarizes this procedure \cite{bouman2020digital}.

In this work, we implement error diffusion halftoning by Floyd-Steinberg dithering \cite{floyd1976an} because of its efficiency and fine-grained results. The Floyd-Steinberg error filter is defined as:
% Floyd-Steinberg error filter
\begin{equation}
\label{hfs}
h_{FS} = \frac{1}{16} \begin{bmatrix} 0 & * & 7 \\ 3 & 5 & 1 \end{bmatrix},	
\end{equation}
where $*$ denotes the pixel being scanned currently, and it only disperses errors to adjacent pixels. The weights are zeros for the pixels that have been scanned, so the error diffusion does not go backward with respect to the raster order.  Alg. \ref{alg} describes this algorithm in detail. For color images, these operations are performed for each channel independently.

\begin{algorithm}[t]
	\label{alg}
	\SetAlgoLined
	\caption{Floyd-Steinberg dithering}
	\KwResult{Output halftone $Q$}
	Given an input image $I$ with pixel values $\in [0,1]$, \\
	\For{$i$ from top to bottom}{
		\For{$j$ from left to right}{		
			$oldValue = I[i][j]$ \\ 
			\eIf{$oldValue > 0.5$}{
				$newValue = 1$ \\
			}{
				$newValue = 0$ \\
			}
			$Q[i][j] = newValue$ \\
			$error = oldValue - newValue$ \\
			$I[i+1][j] \, {+}{=} \, error \times 7 / 16$ \\
			$I[i-1][j+1] \, {+}{=} \, error \times 3 / 16$ \\
			$I[i][j+1] \, {+}{=} \, error \times 5 / 16$ \\
			$I[i+1][j+1] \, {+}{=} \, error \times 1 / 16$ \\
		}
	}
\end{algorithm}

\setlength{\tabcolsep}{9pt}
\renewcommand{\arraystretch}{0.6}
\begin{table*}[htp!]
	\begin{center}
		\caption{Evaluation results (\%) on CIFAR-10. Rows are defense methods, and columns are input types. On the standard training track, models are trained on clean data. On the adversarial training track, the \textit{clean} columns is that models are trained on PGD-$\ell_\infty$ data but tested on clean data. For the other columns, models are trained on a specific attack type (corresponding to the colums).  Avg$_{adv}$ and Avg$_{all}$ denote average accuracies over the four attacks and over all the five data types, respectively.}
		\label{table:results}
		\begin{tabular}{l | c | r | rrrr | rr}
			\hline \noalign{\smallskip} \noalign{\smallskip}
			Method & Training & Clean & PGD-$\ell_\infty$ & PGD-$\ell_2$ & Mult-$\ell_\infty$ & Mult-$\ell_2$ & Avg$_{adv}$ & Avg$_{all}$ \\
			\noalign{\smallskip} \hline \noalign{\smallskip}
			Vanilla & & \textbf{94.03} & 0.01 & 0.20 & 0.05 & 0.01 & 0.07 & 18.86 \\
			Gaussian blur & & \underline{90.17} & 0.20 & 1.34 & 0.17 & 0.05 & 0.44 & 18.39 \\
			Non-local means& Standard & 88.66 & 0.02 & 0.49 & 0.03 & 0.00 & 0.14 & 17.84 \\
			JPEG compression & training & 90.06 & 2.97 & 4.82 & 1.81 & 0.22 & 2.46 & 19.98 \\
			Bit-depth reduction & & 78.87 & \textbf{15.26} & \underline{10.84} & \textbf{10.79} & \textbf{4.52} & \textbf{10.35} & \textbf{24.06} \\
			Halftoning (ours) & & 88.57 & \underline{9.53} & \textbf{11.98} & \underline{5.54} & \underline{1.07} & \underline{7.03} & \underline{23.34} \\
			\noalign{\smallskip} \hline \noalign{\smallskip}
			Vanilla & & \underline{83.31} & \underline{51.15} & \underline{50.68} & 54.10 & 40.29 & \underline{49.06} & \underline{55.91} \\
			Gaussian blur & & 75.96 & 44.59 & 47.12 & 45.07 & 32.48 & 42.32 & 49.04 \\
			Non-local means & Adversarial & 75.47 & 44.67 & 45.29 & 16.59 & 14.53 & 30.27 & 39.31 \\
			JPEG compression & training & 24.97 & 38.99 & 43.72 & \underline{59.15} & \underline{44.72} & 46.65 & 42.31 \\
			Bit-depth reduction & & 71.66 & 47.34 & 42.40 & 48.50 & 41.63 & 44.97 & 50.31 \\
			Halftoning (ours) & & \textbf{84.37} & \textbf{60.01} & \textbf{56.56} & \textbf{67.37} & \textbf{88.44} & \textbf{68.10} & \textbf{71.35} \\	
			\noalign{\smallskip} \hline
		\end{tabular}
	\end{center}
\end{table*}

We deploy Floyd-Steinberg dithering as an image transformation at the pre-processing stage. To elaborate, let $x_{adv}$ be an adversarial example and $T_{FS}$ be the Floyd-Steinberg dithering, then the input of the target model is $T_{FS}(x_{adv})$. Floyd-Steinberg dithering can invalid the adversarial variations of pixel values and destroy the structure of adversarial perturbations through the quantization operation. Moreover, the error diffusion mechanism repeatedly updates the values of the neighboring pixels in the raster order. This makes the adaptive attacks hard to identify the mapping between the original image and the corresponding halftone, so BPDA \cite{obfuscated} would be difficult to approximate the gradients accurately to generate strong adversarial examples. Therefore, Floyd-Steinberg dithering allows us to mitigate adversarial effects in advance, then feed an adversary-free image to the target model for protection. We employ Madry's adversarial training protocol \cite{madry2018towards} to train the model end-to-end. That is, the halftoning transformation is included in the training process so that the model can learn to recognize halftones with adversarial patterns.  On the other hand, spreading quantization errors produces better halftoning quality and tends to enhance edges and object boundary in an image, which are favorable to image recognition tasks. Furthermore, comparing to other complicated error diffusion halftoning algorithms, Floyd-Steinberg dithering only diffuses errors to the adjacent pixels, so it saves computational cost. In short, the proposed method takes adversarial robustness, clean data performance and efficiency into consideration, achieving an excellent balance between these three indicators.

\section{Experiments}

We evaluate our method on four attack types in the white-box setting: PGD-$\ell_\infty$ \cite{madry2018towards}, PGD-$\ell_2$, Mult-$\ell_\infty$ and Mult-$\ell_2$ \cite{lo2020multav}. Mult attack is originally designed for adversarial videos (MultAV), we apply it to generate adversarial images. These attacks include both additive and multiplicative attacks. Clean images are also tested. We compare the performance of our approach with four image transformation-based defenses: Gaussian blur, non-local means \cite{buades2005non}, JPEG compression \cite{dziugaite2016study,das2018shield} and bit-depth reduction \cite{xu2017feature,guo2017countering}. Finally, a deep analysis is provided.

\subsection{Experimental Setup}
%\noindent {\bf{Experimental Setup:}}
We conduct experiments on CIFAR-10 \cite{krizhevsky2009learning}, an image classification dataset that consists of 60,000 images with size $32 \times 32$ from 10 classes. We adopt ResNet-18 \cite{he2016deep} as the backbone network. All the models are trained by the SGD optimizer. We follow Madry's protocal \cite{madry2018towards} for adversarial training.

The settings of the four considered attacks for both inference and adversarial training are described as follows \cite{lo2020multav}. PGD-$\ell_\infty$: $\epsilon=8/255$, $\alpha=3/255$. PGD-$\ell_2$: $\epsilon=1.0$, $\alpha=3.0$. Mult-$\ell_\infty$: $\epsilon_m=1.08$, $\alpha_m=1.03$. Mult-$\ell_2$: $\epsilon_m=1.3$, $\alpha_m=1.03$. $T=5$ for all the attacks.

For the four compared defenses, we set the hyperparameters as follows: The kernel of Gaussian blur is $5 \times 5$ with $\sigma = 1.5$; non-local means is with the Gaussian version; the JPEG compression level is $30/100$; and the bit-depth reduction quantizes pixel values to 1-bit for each channel. Because JPEG compression, bit-depth reduction and halftoning cause obfuscated gradients, we employ BPDA \cite{obfuscated} to mount adaptive attacks for evaluating these defenses. The identity function is used as a surrogate function to approximate the gradients.

\begin{figure}
	\begin{center}
		\includegraphics[width=0.35\textwidth]{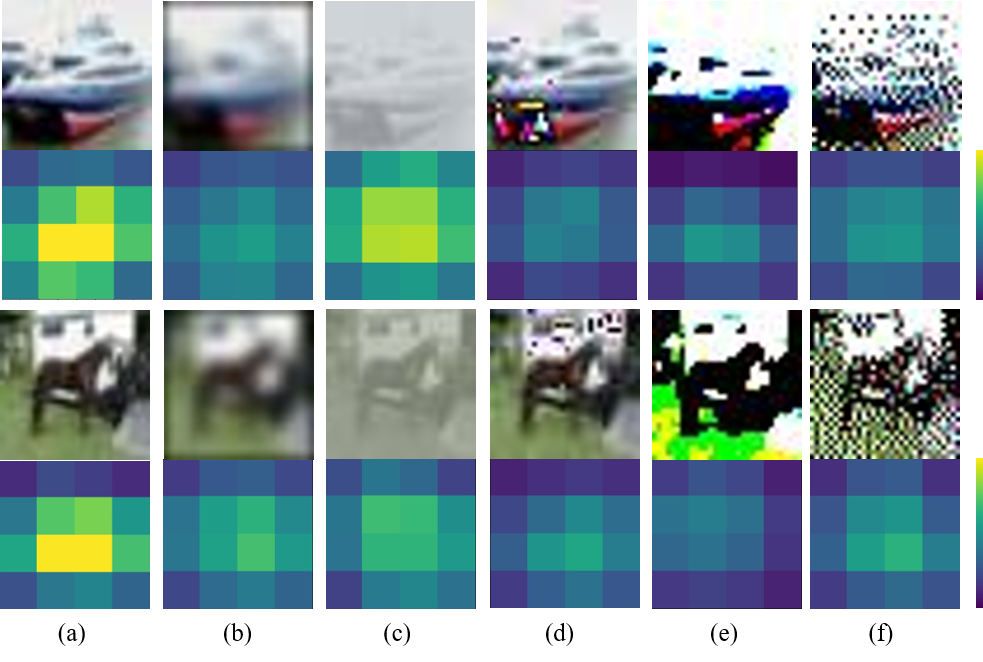}
		\vskip-13pt\caption{Transformed PGD-$\ell_\infty$ adversarial examples, and the visualized differences between the features of clean images and the features of corresponding transformed PGD-$\ell_\infty$ examples. The features (size $8 \times 8$) are from the last conv layer of ResNet-18. (a) Vanilla. (b) Gaussian blur. (c) Non-local means. (d) JPEG compression. (e) Bit-depth reduction. (f) Halftoning. Models are with standard training.}
		\label{fig:visual}
	\end{center}
\end{figure}

\subsection{Evaluation Results}
%\noindent {\bf{Evaluation Results:}}
Table \ref{table:results} reports our experimental results. In the case of standard training, both Gaussian blur and non-local means provides no resistance to any types of adversarial attacks under the white-box threat model. JPEG compression shows somewhat effectiveness. Only bit-depth reduction and halftoning obviously improves the robustness to all of the considered attacks. On the other hand, image transformation-based defenses usually suffer from the drop in clean data accuracy since the transformations degrade the semantic information. Particularly, bit-depth reduction obtains much lower clean data performance. Instead, halftoning is able to enhance the robustness and preserve the performance simultaneously.

Adversarial training has become a backbone of advanced defense approaches. We combine these image transformation-based defense with adversarial training to pursue better robustness. As can be seen from Table \ref{table:results}, adversarial training makes great improvements. However, Gaussian blur, non-local means and bit-depth reduction fail to improve upon the vanilla adversarial training baseline but decrease the performance. Learning features with adversarial patterns is more difficult, so adversarial training requires higher model capability \cite{Xie_2019_CVPR}. These three transformations degrade the model capability and thus cannot handle adversarial training, resulting in worse robustness. Furthermore, Gaussian blur and non-local means are especially vulnerable to Mult attacks, indicating these two image denoising transformations are unable to deal with the multiplicative adversarial perturbations. JPEG compression is useful for Mult attacks but ineffective in PGD attacks. In contrast, the proposed method significantly improves the robustness over all the four considered attacks.

Similarly, adversarial training decreases clean data accuracy \cite{xie2020adversarial}, and the image transformation-based defenses make further degradation. In particular, JPEG compression obtains very low clean data performance when it is trained adversarially. Instead, halftoning's performance drops slightly and achieves the highest clean data accuracy in the adversarial training case, which is even better than the vanilla model. This shows that the halftoning defense is able to handle the difficult adversarial training and can generalize to clean images. These results demonstrate the proposed method is a preferred defense method that can improve robustness against different types of attacks under the white-box setting. Concurrently, it maintains good clean data performance.

\begin{figure}
	\begin{center}
		\includegraphics[width=0.29\textwidth]{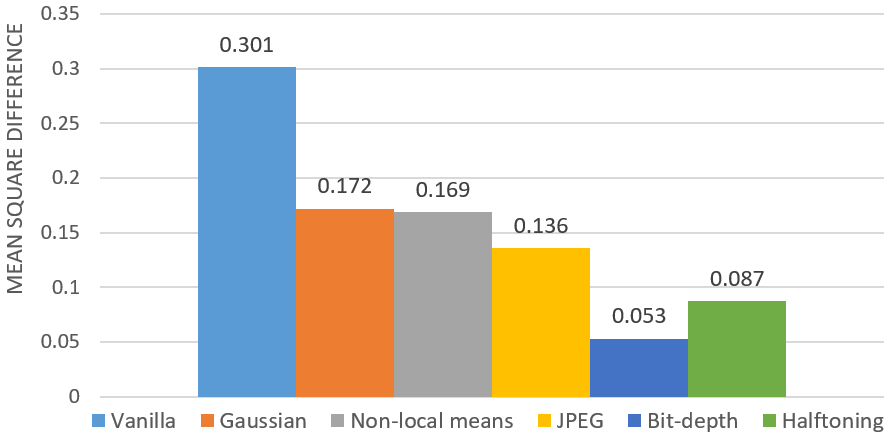}
	\vskip-13pt	\caption{Mean square differences between the features of clean images and the features of corresponding transformed PGD-$\ell_\infty$ examples. The features are from the last conv layer of ResNet-18.  Models are with standard training. The values are the averages of CIFAR-10 test set.}
		\label{fig:chart}
	\end{center}
\end{figure}

\subsection{Analysis}
%\noindent {\bf{Analysis:}}
We display some of PGD-$\ell_\infty$ adversarial examples and their corresponding transformed images in Fig. \ref{fig:visual}. All the transformations lose information to a certain extend, so their clean data performance drops. Particularly, bit-depth reduction produces very coarse images. Instead, halftoning can filter out adversarial perturbations but still maintains highly recognizable image quality.  

Fig. \ref{fig:visual} also shows the visualized differences between the features of clean images and the features of the corresponding transformed PGD-$\ell_\infty$ examples. The quantitative values of such differences are compared in Fig. \ref{fig:chart}. The vanilla model has the largest difference, which means the features are largely changed when the image is adversarially perturbed and thus causes a wrong prediction. Gaussian blur, non-local means and JPEG compression repress the differences yet insufficiently, so their robustness still poor. Bit-depth reduction obtains the smallest difference, but its transformed images are too coarse to recognize accurately. In contrast, halftoning also attains a small difference, showing its features are not easily affected by adversarial perturbation. In the meantime, its image quality is highly recognizable. Hence, our method can achieve good robustness and performance simultaneously.

% To start a new column (but not a new page) and help balance the last-page
% column length use \vfill\pagebreak.
% -------------------------------------------------------------------------
%\vfill
%\pagebreak

\section{Conclusion}

In this paper, we propose a novel image transformation-based defense method by using Floyd-Steinberg halftoning. The 1-bit quantization and error diffusion mechanisms can remove adversarial perturbations and weaken adaptive attacks. Furthermore, the proposed method's ability to produce high-quality halftones guarantees good clean data performance. Although the majority of the image transformation-based defenses have been shown to be ineffective under the white-box threat model, our method is able to greatly improve adversarial robustness. We show that this defense stream is still promising and worthy to explore.

% References should be produced using the bibtex program from suitable
% BiBTeX files (here: strings, refs, manuals). The IEEEbib.bst bibliography
% style file from IEEE produces unsorted bibliography list.
% -------------------------------------------------------------------------

\bibliographystyle{IEEEbibAbb}
\bibliography{cite}

\end{document}